\documentclass[twoside,11pt]{article}

\usepackage{jmlr2e}
\usepackage[utf8]{inputenc} 
\usepackage[T1]{fontenc}    
\usepackage{hyperref}       
\usepackage{url}            
\usepackage{booktabs}       
\usepackage{amsfonts}       
\usepackage{nicefrac}       
\usepackage{microtype}      
\usepackage{xcolor}         
\usepackage{amsmath}
\usepackage{graphicx}
\usepackage[ruled,vlined]{algorithm2e}
\usepackage{adjustbox}
\usepackage{subcaption}
\usepackage{wrapfig,lipsum,booktabs}
\usepackage{graphicx}
\usepackage{selectp}
\usepackage{multirow}
\usepackage{enumitem}

\makeatletter
\def\section{\@startsection{section}{1}{\z@}{-0.09in}{0.02in}{\large\bf\raggedright}}
\def\subsection{\@startsection{subsection}{2}{\z@}{-0.08in}{0.01in}{\normalsize\bf\raggedright}}
\def\subsubsection{\@startsection{subsubsection}{3}{\z@}{-0.07in}{0.01in}{\normalsize\sc\raggedright}}
\def\paragraph{\@startsection{paragraph}{4}{\z@}{0.5ex plus 0.5ex minus .2ex}{-0.5em}{\normalsize\bf}}
\def\subparagraph{\@startsection{subparagraph}{5}{\z@}{1.5ex plus 0.5ex minus .2ex}{-1em}{\normalsize\bf}}
\makeatother
\jmlrheading{1}{}{}{}{}{}

\firstpageno{1}

\title{Probabilistic Generative Modeling for\\ Procedural Roundabout Generation for Developing Countries}
\begin{document}


\author{\name Zarif Ikram \email 1905111@ugrad.cse.buet.ac.bd \\
       \addr Department of Computer Science and Engineering\\
       Bangladesh University of Engineering and Technology, Dhaka, Bangladesh
       \AND
       \name Ling Pan  \email penny.ling.pan@gmail.com\\
       \addr Mila - Qu\'ebec AI Institute
       \AND
       \name Dianbo Liu \email dianbo.liu@alumni.harvard.edu\\ 
       \addr  Mila - Qu\'ebec AI Institute and NUS, Singapore }


\maketitle

\begin{abstract}

Due to limited resources and fast economic growth, designing optimal transportation road networks with traffic simulation and validation in a cost-effective manner is vital for developing countries, where extensive manual testing is expensive and often infeasible. Current rule-based road design generators lack diversity, a key feature for design robustness. Generative Flow Networks (GFlowNets) learn stochastic policies to sample from an unnormalized reward distribution, thus generating high-quality solutions while preserving their diversity. In this work, we formulate the problem of linking incident roads to the circular junction of a roundabout by a Markov decision process, and we leverage  GFlowNets as the Junction-Art road generator. We compare our method with related  methods and our empirical results show that our method achieves better diversity while preserving a high validity score.

\end{abstract}


\section{Introduction}
\label{intro}
The road transportation system plays a pivotal role in a nation's infrastructure development, as it fosters economic improvement by enhancing transportation. Effective planning and implementation of road designs are critical, as they not only shape and transform a country's landscape but also lay the foundation for potential adaptation for autonomous vehicles in the future. Nevertheless, in many developing countries, resource constraints impede their capacity to invest extensively in optimal road design. Extensive manual road design is often impractical due to the cost and time required, making road generators a more practical alternative in the real world. 

Road design diversity is an important aspect of these generators, as it enables road designers to explore a wider range of options for optimal development. Recent advancements in road generators have enabled the creation of complex and diverse roads. 
However, many of these generators do not adequately address several intricate road structures commonly found in the real world.
Among these structures, the generation of diverse roundabouts is particularly crucial, as they are one of the most common road configurations encountered in reality.  Additionally, since roundabouts serve as modular components within other road networks, they need to exhibit diversity within a specific incident road network, providing multiple options to enhance flexibility and adaptability in road design and planning. Consequently, both diversity and realism are equally important in the context of road design.
\begin{figure}[ht]
\centering
    \begin{subfigure}[b]{.25\linewidth}
      \centering
      \includegraphics[width=.95\linewidth]{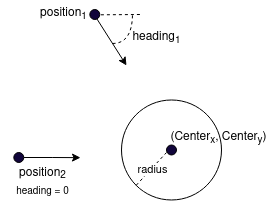}
      \caption{Example problem}
      \label{fig:problem}
    \end{subfigure}
    \begin{subfigure}[b]{.5\linewidth}
      \centering
      \includegraphics[width=.95\linewidth]{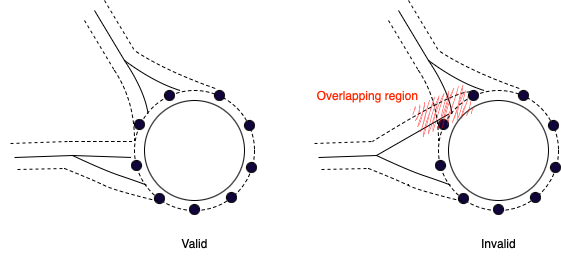}
      \caption{Solution}
      \label{fig:solution}
    \end{subfigure}
    \vspace{-.1in}
    \caption{For any given incident road configurations, incident roads can connect to the circular roads in many valid ways.}
    \label{fig:valid}
\end{figure}

\vspace{-.1in}
In this paper, we propose to automatically generate such roundabouts with Generative Flow Networks (GFlowNets)~\citep{Bengio2021GFlowNetF}, which is a generative model that can sample roundabouts proportional to a reward function. By employing this method, we can sample and discover roundabouts that cover the different modes of the reward function specified for realism. 
In this work, we formulate the problem of  connecting incident roads in the circular roads of the roundabout by a Markov decision process and connect GFlowNets to the Junction-Art (\cite{JuncArt1}) road generator to sample diverse roundabouts. To the best of the authors' knowledge, this is the first work that incorporates GFlowNets for road generation. The key contributions are the following.

\begin{itemize}[leftmargin=*]
    \item A Markov Decision Process(MDP) formulation for connecting incident roads to the circular road junction of a roundabout
    \item A novel algorithm based on GFlowNets for diverse roundabout configuration generation with a carefully designed diversity measure in the context of road generation
    \item Strong empirical experimental results in terms of diversity and performance in roundabout generation
\end{itemize}

\section{Problem Setup} \label{sec:setup}

The $r$ road configurations are denoted by $P = \{(\text{position}_1,\text{heading}_1, n_{\text{leftlanes},1}, n_{\text{rightlanes},1}), \dots, \\(\text{position}_r, \text{heading}_r, \allowbreak n_{\text{leftlanes},r}, n_{\text{rightlanes},r})\}$, 
where each tuple represents the position, heading of the starting point of an incident road and the number of incoming and outgoing lanes it has.
We want to generate a roundabout by creating the circular road configuration with $(\text{Center}_x, \text{Center}_y, radius)$, where $(Center_x, Center_y)$ represents the center coordinate and $radius$ represents the radius of the circular shape of the roundabout. We build the circular road with $C$ parametric cubic road segments and connect each N lanes $N = \sum_{i \in \{1, \dots, r\}}{n_{leftlanes,i} + n_{rightlanes,i}}$ to one of the $C$ circular road segments ${s_1, \dots, s_C}$. We wish to create a roundabout for which the configuration of the connection is valid. Formally, we want to search over a (terminal) state space of $\mathcal{X}$ to find configurations $x \in \mathcal{X}$  
that have a validity reward $f : x \rightarrow R^+$ which indicates design and construction feasibility of the roundabout where $x = \{s_{(x_1)}, \dots, s_{(x_N)}\}$, while maintaining high diversity. For example, Figure \ref{fig:valid} represents a problem where $C = 9$ and $N = 4$, and Figure \ref{fig:mdp} represents a problem where $C = 10$, $N = 4$, and $x = \{8, 6, 1, 9\}.$

In this context of this problem, we follow the algorithm proposed in \cite{zarifRoundabout} to generate circular road segments. We want to generate diverse configurations for which the generated roundabouts are valid, i.e., there are no overlaps between any incident roads and circular road segments. Finally, we generate diverse roundabouts by connecting the incident lanes with circular segments according to the configuration.

\subsection{Criteria for Roundabout Diversity}
As the main goal of this work is the diversification of the roundabouts, we present a measure of diversity of the candidates. We adopt the diversity measure proposed in \cite{biological} for our work, i.e., for a set of generated candidates $\mathcal{D}$,

\begin{equation}
\label{eq:diversity}
    Diversity(\mathcal{D}) = \frac{\sum_{x_i \in \mathcal{D}}{\sum_{x_j \in \mathcal{D}\setminus\{x_i\}}{d(x_i, x_j)}}}{|\mathcal{D}|(|\mathcal{D}| - 1)}
\end{equation}

where $d$ is the distance function. In our work, the diversity of the generated roundabouts emerges from the diversity of the connection roads between the incident and circular roads. As a result, we define the distance function $d$ as the mean discrete Fr{\'e}chet distance (\cite{frechet}) between each connection road geometries. To discretize each geometry, we sample $m$ equidistant coordinates from the geometry where $m$ = 10.
\section{Method}
\begin{figure}[bt!]
    \centering
    \includegraphics[width=.7\linewidth]{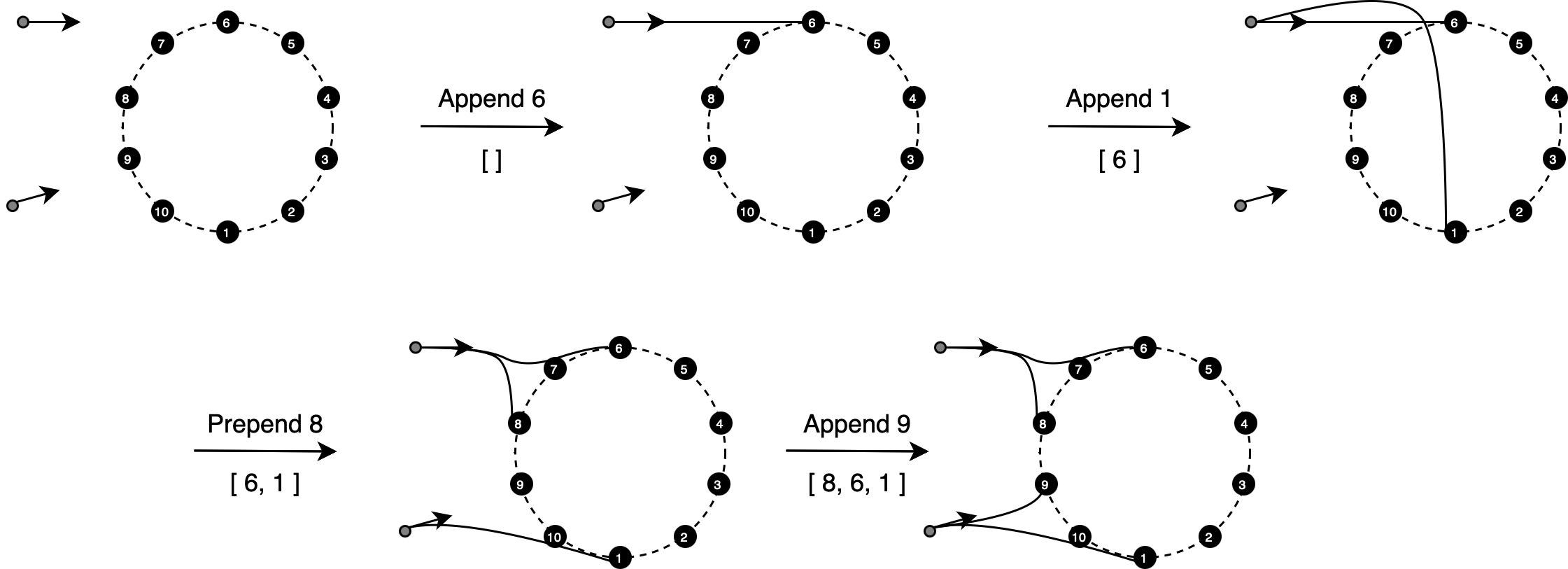}
    \caption{MDP design for incident road configuration generation. Number on dark filled circles on the dotted circle represent the endpoints of circle road segments. Small filled circle with arrow represent the road definition where direction of the arrow represent the heading of the initial road.}
    \label{fig:mdp}
\end{figure}
\subsection{GFlowNets}
GFlowNets (\cite{Bengio2021FlowNB}, \cite{Bengio2021GFlowNetF}) are amortized samplers that can sample from high-dimensional state spaces. Given a state space $\mathcal{X}$, they can learn a stochastic policy $\pi$ that can sample states $x \in \mathcal{X}$ using a non-negative reward function $R(x)$ such that $\pi(x) \propto R(x)$. This property of GFlowNets makes them the ideal candidate for generating diverse yet valid roundabouts.

We construct objects $x \in \mathcal{X}$ by sampling constructive, irreversible actions $a \in \mathcal{A}$ that transition $s_t$ to $s_{t+1}$. Therefore, in  graph terminology, GFlowNets generate a graph $(\mathcal{S}, \mathcal{E})$ where each node in $\mathcal{S}$ denotes a state in the Markov chain with an initial state $s_0$, and each edge in $\mathcal{E}$ denotes a transition $s_t \rightarrow s_{t+1}$ with a special terminal action indicating $s = x \in \mathcal{X}$. Moreover, the graph is an acyclic DAG. A trajectory $\tau \in \mathcal{T}$is a Markov chain $(s_0\rightarrow s_1\xrightarrow{}\dots \rightarrow x), x\in\mathcal{X} $ where $\mathcal{T}$ is the set of all trajectories.  

By learning the stochastic forward and backward policies using  a neural network, GFlownets learn to construct objects with the aforementioned properties. The forward policy $P_F(s_{t+1}|s_t; \theta)$ represents the probability distribution from $s_t$ to $s_{t+1}$, where the neural network is parameterized by $\theta$. The backward policy $P_B$ can be chosen freely, and here we use a uniform $P_B$, which is uniformly distributed among allowed actions from $s_t+1$ to $s_t$. Therefore, the probability of sampling an object $x\in\mathcal{X}$ is:
\begin{equation}
    \pi(x) = \sum_{\tau\in\mathcal{T}:x\in\tau} \prod_{t=0}^{|\tau-2|}{P_F(s_{t+1}|s_{t};\theta)} 
\end{equation}

We use the trajectory balance objective (\cite{trajectory}) to train the GFlowNet, which is defined as in Eq.~(\ref{lossFunction}), where $Z_\theta$ is a learnable parameter that represents the total flow: $\sum_{x\in\mathcal{X}}{R(x)} =  \sum_{s:s_0\rightarrow s\in\tau \forall \tau \in \mathcal{T}}{P_F(s|s_0;\theta)}$.

\vspace{-.1in}
\begin{equation}  
\label{lossFunction}
    \mathcal{L}(\tau; \theta) = \big(\log Z_\theta + \log \sum_{s_t\rightarrow s_{t+1} \in \tau}{P_F(s_{t+1}|s_t; \theta)} - \log R(x) - \log \sum_{s_t\rightarrow s_{t+1}\in\tau}{P_B(s_t, s_{t+1};\theta)} \big)^2
\end{equation}



\subsection{MDP For Incident Lane Connection Configuration}
Here, we define the actions required to generate the MDP in the context of our problem. Starting from an empty state {}, we append or prepend a slot number $s_i, i \in {1, \dots, C}$ to the state. We, then, conclude the MDP with a special $terminate$ action. Therefore, the action space is defined as
$\mathcal{A} = \{a_1, \dots, a_C, p_1, \dots, p_C, TERMINATE \}$.
Here $a_i$ and $p_i$ represent the actions of appending and prepending slot $i$ to the current state, respectively. This choice of action space makes for a constant trajectory ($|\mathcal{T}| = N + 1$) while keeping the action space relatively small. Figure \ref{fig:mdp} shows a detailed example of the generation of a valid configuration using this MDP.

\begin{figure}[t]
    \centering
    \begin{subfigure}[b]{.3\linewidth}
      \centering
      \includegraphics[width=\linewidth]{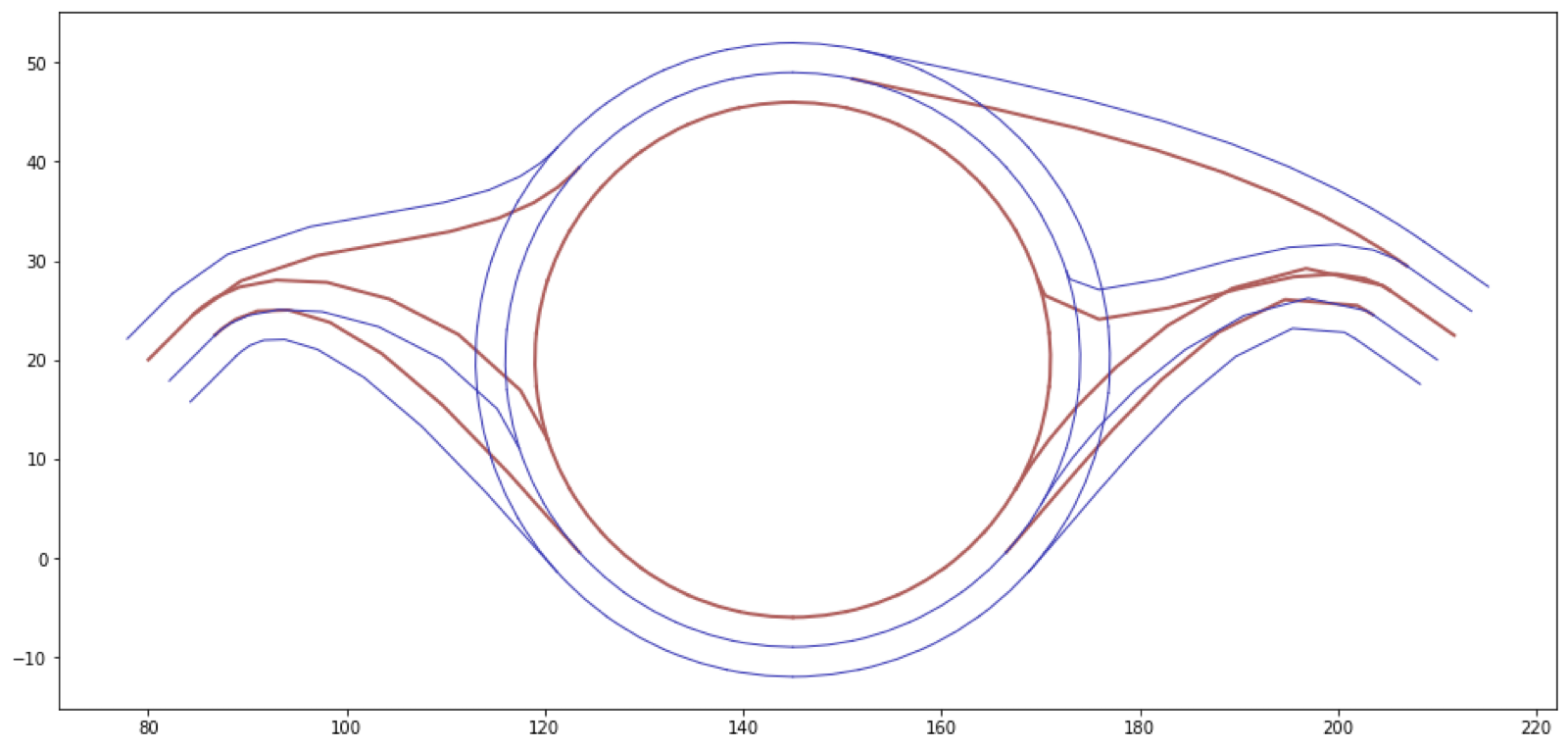}
      \label{real}
    \end{subfigure}
    \begin{subfigure}[b]{.35\linewidth}
      \centering
      \includegraphics[width=\linewidth]{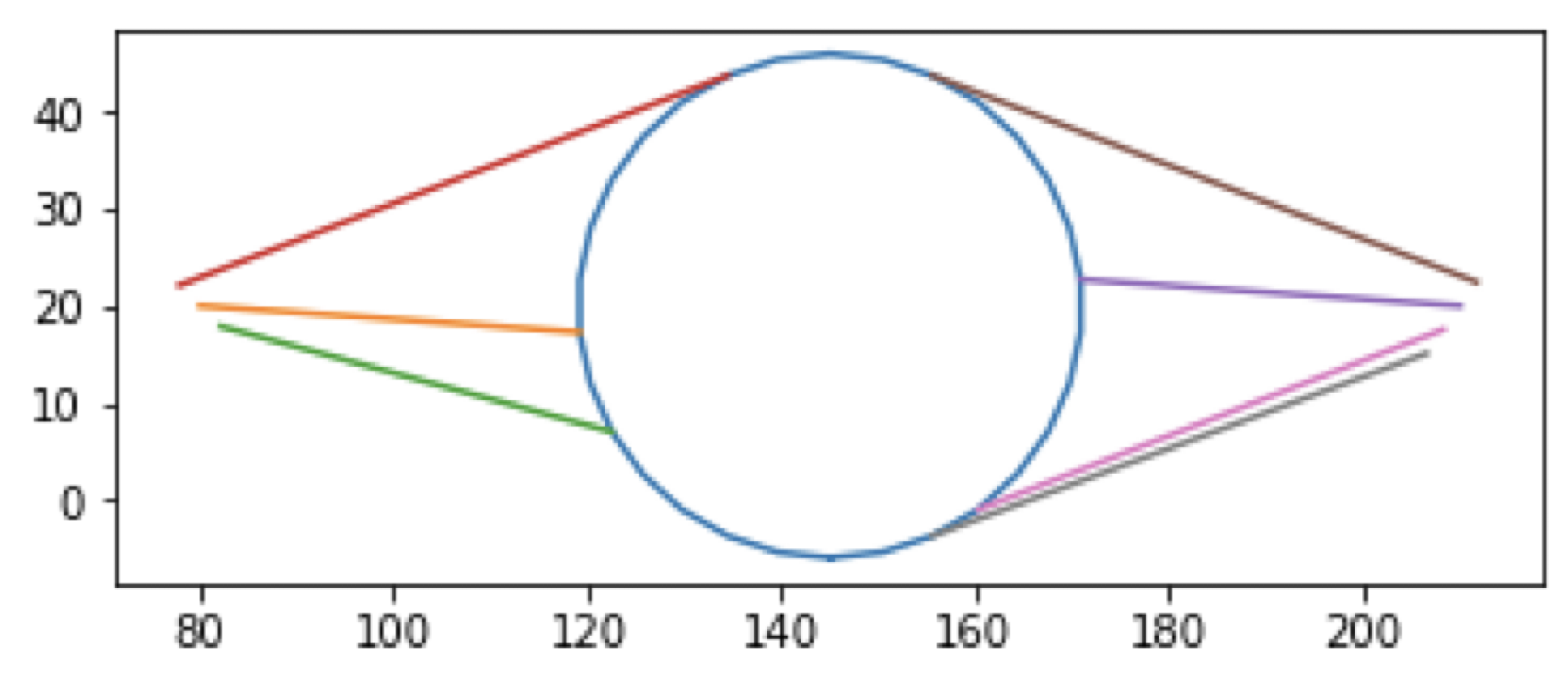}
      \label{proxy}
    \end{subfigure}
      \vspace{-.1in}
  \caption{\textit{Left}: Geometry of a generated roundabout. \textit{Right}: Geometry generated by the proxy reward function.}
    \vspace{-.1in}
  \label{fig:proxyVsReal}
\end{figure} 

\subsection{GFlowNets for Roundabout Design}
We begin by noting that the search space is $C^N$, which scales rapidly with N. To make matters worse, roundabout validation is costly, as each roundabout needs to be generated first to validate, which is time-intensive. On a desktop, one roundabout takes 200ms to generate, meaning we can validate only 18000 configurations in an hour, making GFlowNet training impractical with a reward function that depends on proper validation of generated configurations. To tackle this challenge, given $f(line_i, line_j)$ returns 1 or 0 depending on whether $line_i$ and $line_j$ intersects or not, we propose a proxy reward function that approximately validates the configuration by associating it with the following reward.


\begin{equation}
\label{rewardFunction}
    D(x) = \sum_{i < j}^{i, j \in N} \left(1 - f(line_i, line_j)\right) + \sum_i^{i \in N} (1 - f(line_i, circle))\\
\end{equation}
\vspace{-.1in}
\begin{equation}
\label{eq:rewardFunctionNormalized}
\begin{split}
    D_{normalized}(x) = &\frac{D(x)}{\frac{N(N - 1)}{2} + N}\\
\end{split}
\end{equation}
\begin{equation}
    \label{eq:proxyReward}
    R_{proxy}(x) = base^{D_{normalized}(x)}
\end{equation}

We illustrate the workings of the proxy reward function using Figure \ref{fig:proxyVsReal}. The proxy reward function creates straight lines $line_i$ between the point associated with $s_{(x_i)}$ and the endpoint of $lane_i$ $\forall i \in \{1, \dots, N\}$. Using straight lines, we can approximately validate the configuration by associating a reward with each non-intersecting lines and circle. Besides, to sample from the high-reward region only, we use a reward exponent.

We train the GFlowNet for a specific input of $N$, $C$, and $P$. For each episode, we begin by generating $m$-sized minibatches of trajectories $\mathcal{T}$. Notably, we select actions by sampling using $P_F$ from valid actions from $s$. Upon receiving $TERMINATE$, we calculate the proxy reward from $\mathcal{T}$ and forward and backward flows. Finally, we calculate $\mathcal{L}$ and update parameters $\theta$. We detail the process in algorithm \ref{Algo:train} in Appendix \ref{app:algorithm}.

For roundabout generation, we sample batch $B$ according to GFlowNet policy $\pi_\theta$ and extract $K$ configurations with the highest proxy reward. Initially, we create circular junctions using $P$ and $C$, followed by the incident roads. Using $top-K$, we connect the circular junctions using cubic polymetric roads and evaluate the validity of the connections. Finally, we extract the distinct samples that generate valid roundabouts. Algorithm \ref{Algo:generate} from Appendix \ref{app:algorithm} shows the procedure.

\section{Experiments}
\subsection{Results \& Analysis}
\paragraph{Experimental Setup} We now analyze the effectiveness of our method for roundabout generation as detailed in Section~\ref{sec:setup}.
We follow \cite{zarifRoundabout} for generating input road configurations, which generates a random circle, followed by sampling points from the circle and adding noise to the sampled coordinates and their headings. We generate 20 road configurations for $N = 4, 6, $ and $8$. To ensure comparability, we use the same input for all methods. We compare our method with \cite{zarifRoundabout} and a strong reinforcement learning method --- Soft Actor-Critic (SAC) (\cite{sac}) for comparison which is widely used to generate diverse states by probabilistically sampling the action space. We use a discrete SAC (\cite{sacDiscrete}) with a 64-dimentional hidden layer for both actor and critic networks, a replay buffer of 5000 capacity, as well as $\gamma = 0.99$, $\tau = 0.01$, and a parameterized $\alpha$. We construct the GFlowNet agent with a 128-dimensional hidden layer for the forward layer and a uniform distribution for the backward layer. Finally, we train the GFlowNet and SAC agents for 1000 and 2000 iterations, respectively. 
\paragraph{Analysis} We begin the evaluation by sampling $10^4$ samples, selecting top-K samples based on proxy rewards, and generating roundabouts from them. Then, we calculate the scores based on the number of non-overlapping lanes and the diversity based on Equation \ref{eq:diversity}. We detail the results in Table \ref{table-comparison}. Our analysis show that both GFlowNets and SAC achieves comparable scores to the baseline. Due to maximizing proxy rewards, SAC achieves higher score than GFlowNets in small tasks $(N = 4)$, but lacks diversity, falling short in case of bigger tasks or higher sampling.

\begin{table}[!htb]
  \caption{Roundabout generation experimental results. Baseline uses rule-based method, and so cannot generate diverse roundabouts. Thus, "---" denotes its diversity.}
  \label{table-comparison}
  \centering
  \begin{tabular}{cccccccc}
    \toprule
          \multirow{2}{*}{$N$} & \multirow{2}{*}{METHOD}  & \multicolumn{3}{c}{$K$ = 50} & \multicolumn{3}{c}{$K$ = 200}\\
         \cmidrule{3-5}  \cmidrule{6-8}
         & & & SCORE & DIVERSITY & & SCORE & DIVERSITY  \\
 \midrule
 \midrule
    \multirow{3}{*}{$4$} & BASELINE & & 6.6 $\pm$ 1.3 & --- & & 6.6 $\pm$ 1.3 & ---  \\
    & SAC  & & 6.8 $\pm$ 0.6 & 0.4 $\pm$ 0.5 & & 6.7 $\pm$ 0.8 & 0.5 $\pm$ 0.8 \\
    & OURS & & 6.6 $\pm$ 0.7 & 4.7 $\pm$ 1.7 & & 6.6 $\pm$ 0.7 & 4.8 $\pm$ 1.5  \\
    \midrule
    \multirow{3}{*}{$6$} & BASELINE & & 17.4 $\pm$ 0.8 & --- & & 17.4 $\pm$ 0.8 & ---  \\
    & SAC  & & 15.0 $\pm$ 1.1 & 2.4 $\pm$ 1.5 & & 14.8 $\pm$ 1.2 & 2.9 $\pm$  1.7 \\
    & OURS & & 15.6 $\pm$ 0.9 & 6.3 $\pm$ 0.6 & & 15.6 $\pm$ 1.0 & 7.0 $\pm$ 0.5  \\
    \midrule
    \multirow{3}{*}{$8$} & BASELINE & & 30.3 $\pm$ 1.2 & --- & & 30.3 $\pm$ 1.2 & ---  \\
    & SAC  & & 25.8 $\pm$ 2.1 & 3.4 $\pm$ 2.0 & & 25.1 $\pm$ 2.3 & 3.4 $\pm$ 2.0  \\
    & OURS & & 28.1 $\pm$ 1.4 & 7.0 $\pm$ 0.48 & & 27.8 $\pm$ 1.5 & 7.9 $\pm$  0.4 \\
    \bottomrule
  \end{tabular}
\end{table}


\subsection{Ablation Study}
\begin{figure}[!h]
\centering
    \begin{subfigure}[b]{.33\linewidth}
      \centering
      \includegraphics[width=.95\linewidth]{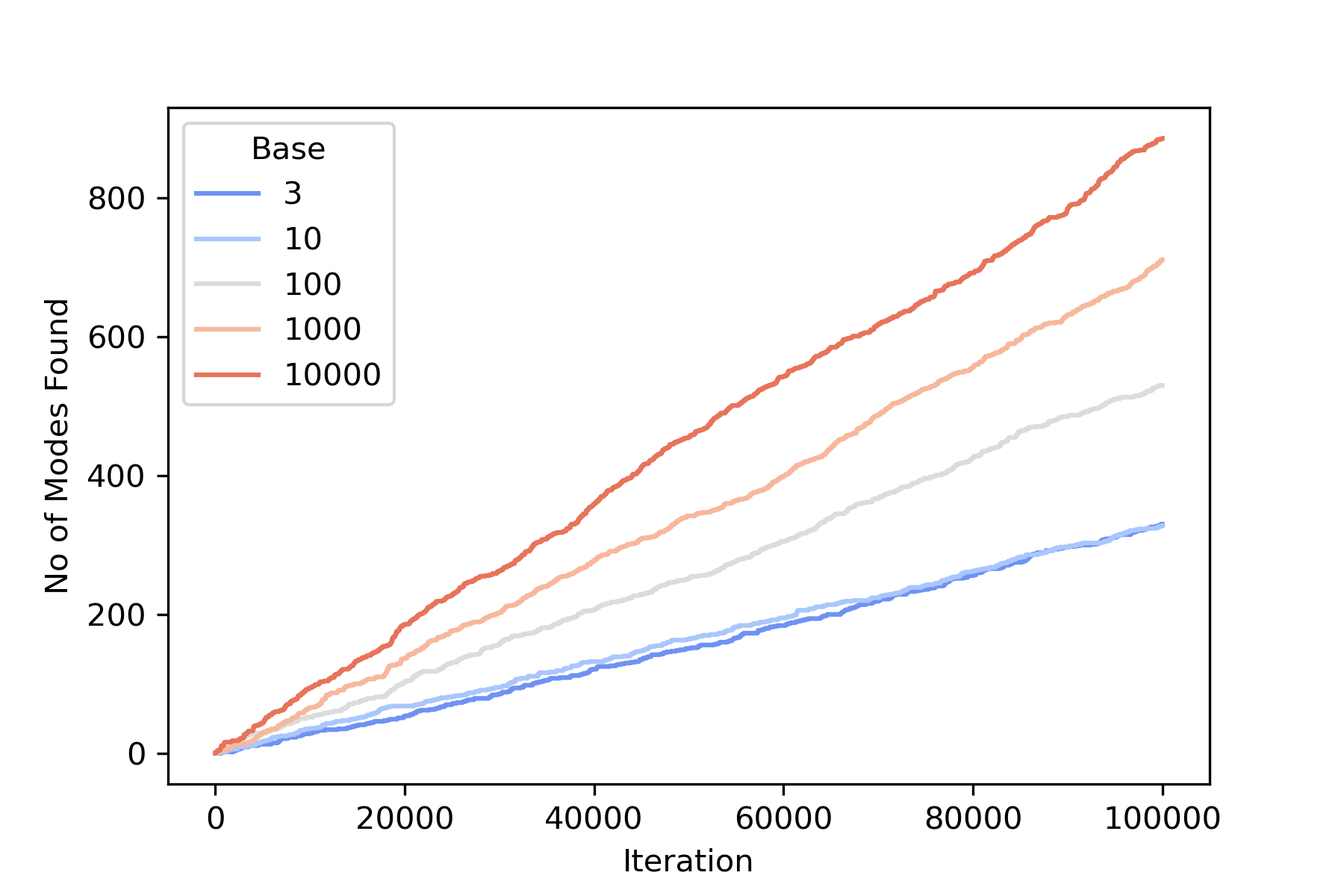}
      \caption{}
      \label{fig:base}
    \end{subfigure}
    \begin{subfigure}[b]{.33\linewidth}
      \centering
      \includegraphics[width=.95\linewidth]{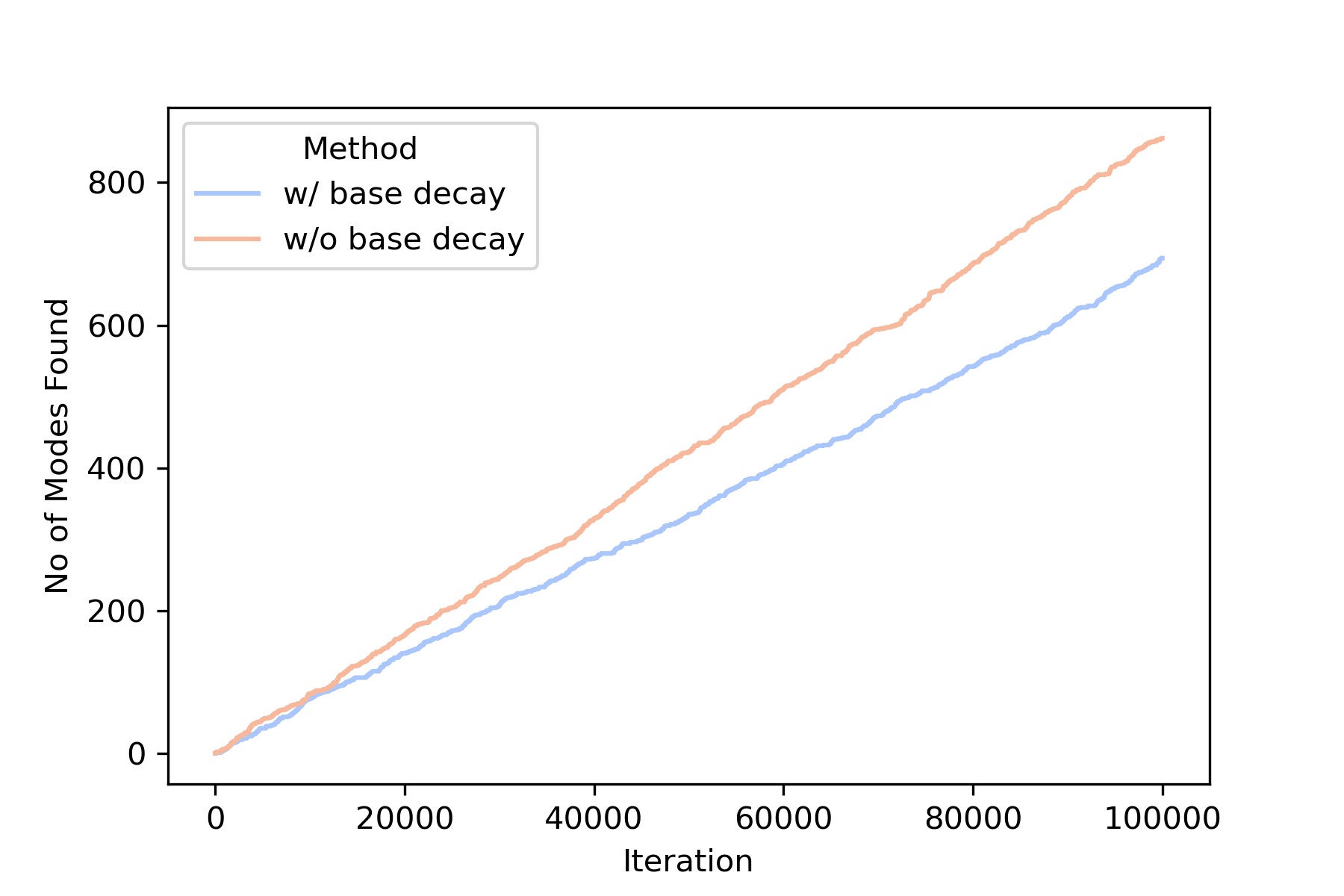}
      \caption{}
      \label{fig:baseDecay}
    \end{subfigure}
    \begin{subfigure}[b]{.33\linewidth}
      \centering
      \includegraphics[width=.95\linewidth]{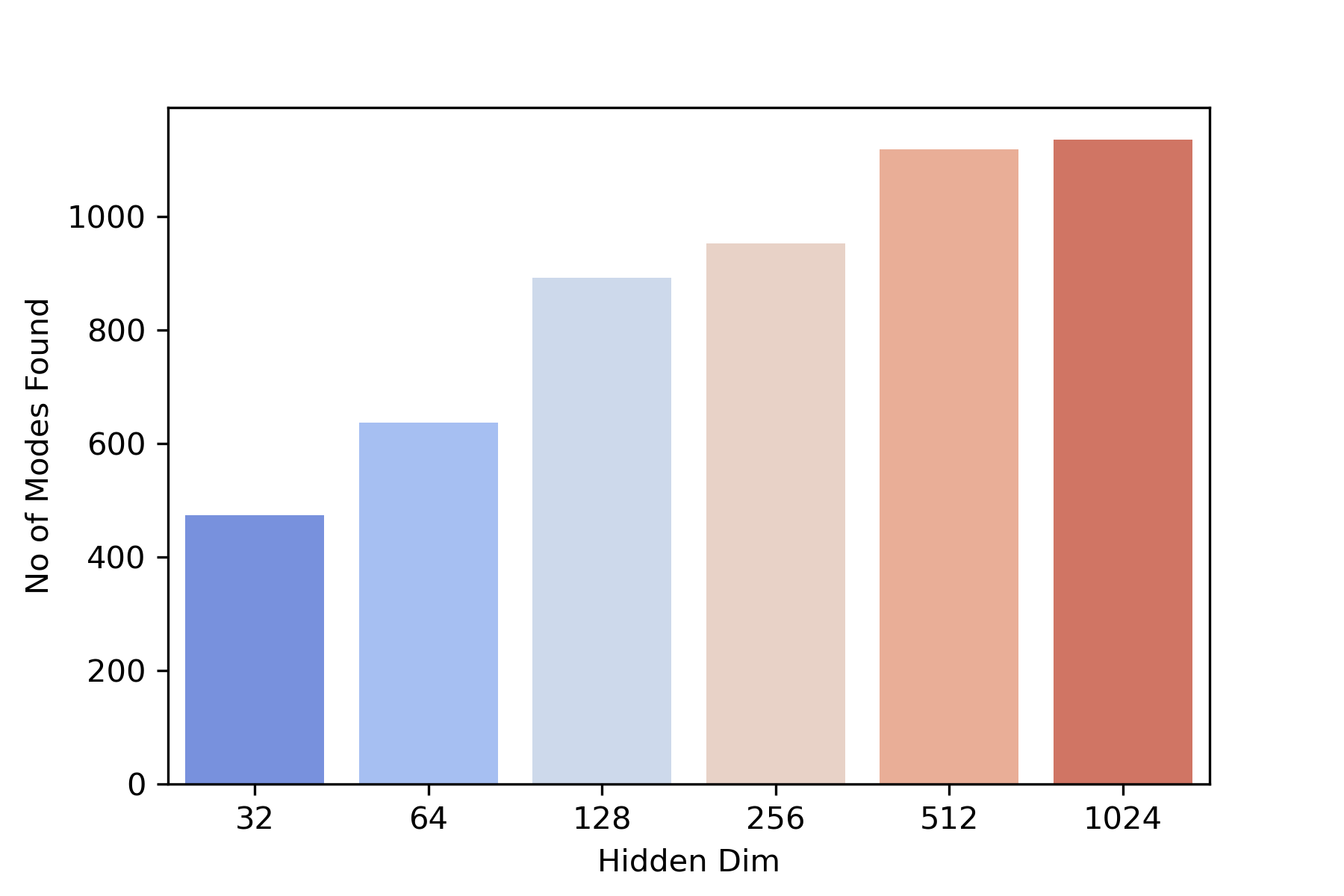}
      \caption{}
      \label{fig:hidden}
    \end{subfigure}
    \vspace{-.1in}
    \caption{Ablation study on GFlowNet training. }
    \label{fig:ablationStudy}
\end{figure}

To evaluate the design choices for our model, we conduct an in-depth ablation study next. First, we vary $base$ from proxy reward $R_{proxy}$ in Equation \ref{eq:proxyReward} and observe the number of diverse modes sampled from each model. For the ablation study, we define a mode for which $D_{normalized}$ from Equation \ref{eq:rewardFunctionNormalized} is more than 0.85. We observe a positive correlation with the base and number of sampled modes. Furthermore, to experiment with the effect of decaying the base in training, we train two models with and without base decay. Figure \ref{fig:baseDecay} shows that decaying base, in fact, results in a smaller number of modes sampled. We also observe a trend with the number of hidden parameters and sampled modes, although the effect diminishes with increasing parameters (Figure \ref{fig:hidden}). 
\vspace{-.1in}
\section{Conclusion}
Automatic road generation is important for intricate road design in developing nations for its cost-effectiveness. Diverse road generation can pave the way for better infrastructure development by providing designers and builders with more options. Roundabouts are notoriously underrepresented in road generators for their difficult intersection design, despite being an important component in road design. As the pluggable component within a set of already established road networks, generated roundabouts need to be diverse for the same set of incident roads. Based on Junction-Art, a high-definition (HD)-map generator, we provide a novel algorithm based on GFlowNets that can sample valid yet diverse configurations for roundabout generation. We also show that the roundabouts discovered by our algorithm are novel and diverse by comparing our method with other methods.



\vskip 0.2in
\bibliography{References}
\newpage

\appendix

\begin{centering}
\textit{\large Supplementary material for}

\textbf{\large Probabilistic Generative Modeling for \\Procedural
Roundabout Generation for Developing Countries}

Zarif Ikram, Ling Pan \& Dianbo Liu

\end{centering}
\section{Related Work}
\subsection{Automatic Road Generation}

Due to the importance of HD maps in AV simulation testing, many works tackle the problem of road generation, most of which can be divided into two groups: (a) from real-world data (\cite{streetgen}, \cite{mapGenerationExtractFeature}, \cite{highfidelity}, \cite{chen2008interactive}) and (b) from constructive generation (\cite{ASFault}, \cite{PCGOSM}, \cite{pcg1}, \cite{zarifRoundabout}). Recently, more advanced work such as Junction-Art \citep{JuncArt1} proposes using procedural content generation (PCG) to produce road maps that are directly pluggable to existing simulation environments such as CARLA \citep{CARLA} using OpenDrive format. \cite{Intersection} brings intersection generation to that work. However, current generation of road-generators lacked roundabout generation, which, more recently, \cite{zarifRoundabout} addressed, but it generated roundabouts using a rule-based method, producing a fixed set of roundabouts. As discussed in Section \ref{intro}, we improve that aspect using GFlowNets sampling in this work.

\subsection{GFlowNets for Generation}
Many recent works show employing GFlowNets diverse candiate generation, such as for biological sequence design \citep{biological}, scientific discovery \citep{aiDriven}, terrain generation \citep{terrainGeneration}, and game level design \citep{gameLevel}. Notably, given the methods used, \cite{gameLevel} is the closest to our work.

\subsection{Reinforcement Learning Methods for Discrete Action Space}
Many reinforcement larning (RL) algorithms such as \cite{dqn} and \cite{ppo} promise proxy reward optimization for discrete action spaces. Soft actor critic algorithms (\cite{sac}) propose improved exploration by optimizing both its rewards as well as the entropy. Albeit originally proposed for continuous action space, the algorithm finds its application in discrete action space as well (\cite{sacDiscrete}, \cite{sacDiscrete2}).

\section{Other Experiments}
\label{app:otherExperiments}
\begin{figure}[h]
    \centering
    \includegraphics[width=.9\linewidth]{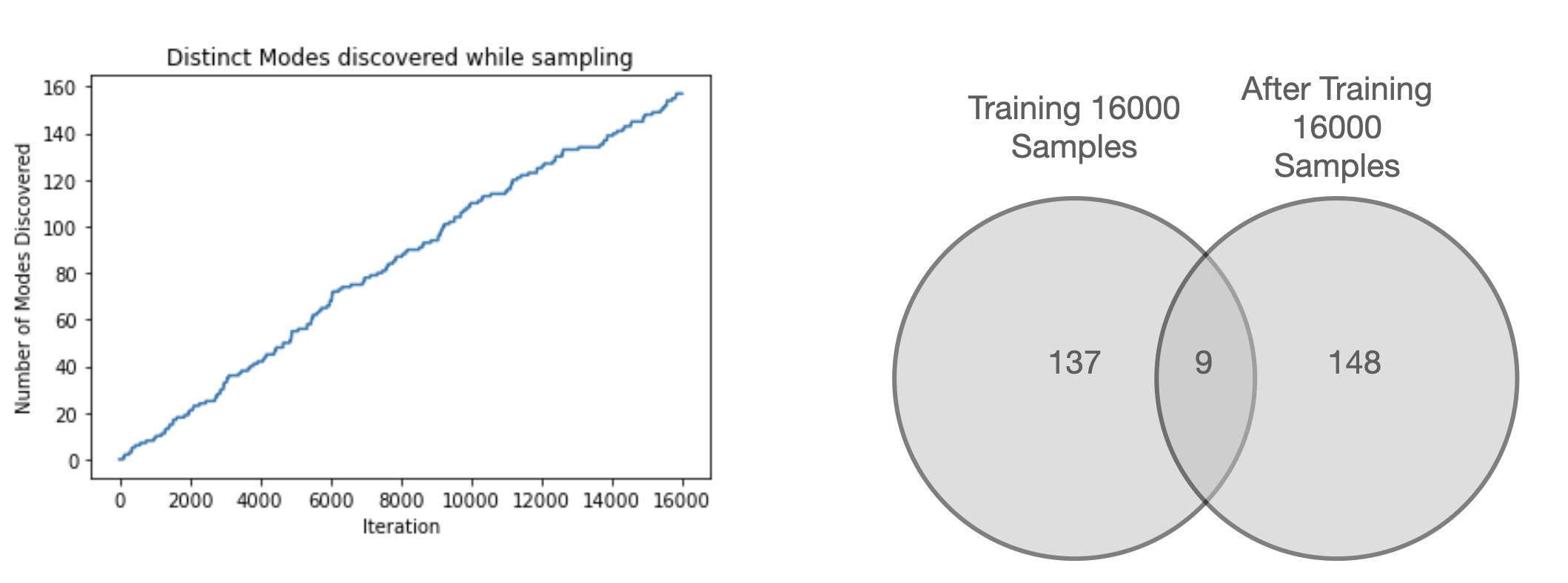}
    \caption{Left diagram shows the number of distinct modes 
    discovered as a function of number of samples for $N$ = 5 and $C$ = 20. Right diagram shows that the agent discoverers configurations not seen previously in training.}
    \label{fig:fiveLanes}
\end{figure}

\begin{figure}[h]
    \centering
    \begin{subfigure}[b]{.35\linewidth}
      \centering
      \includegraphics[width=.95\linewidth]{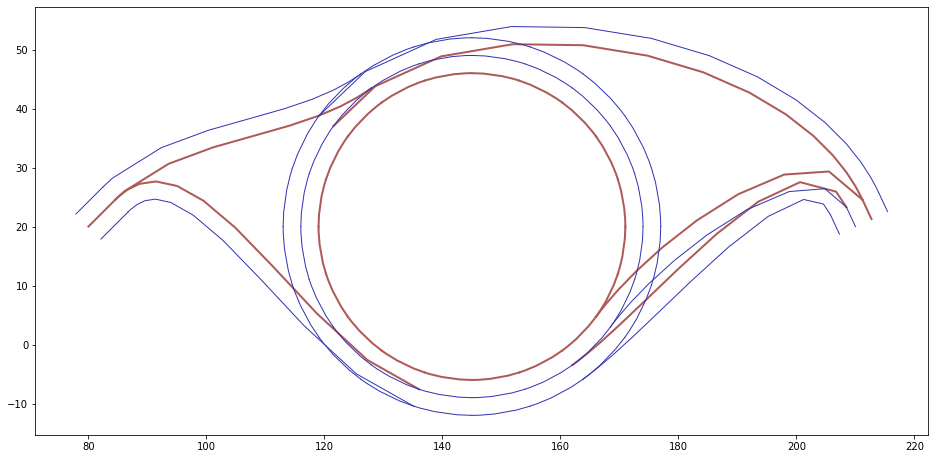}
    
    \end{subfigure}
    \begin{subfigure}[b]{.35\linewidth}
      \centering
      \includegraphics[width=.95\linewidth]{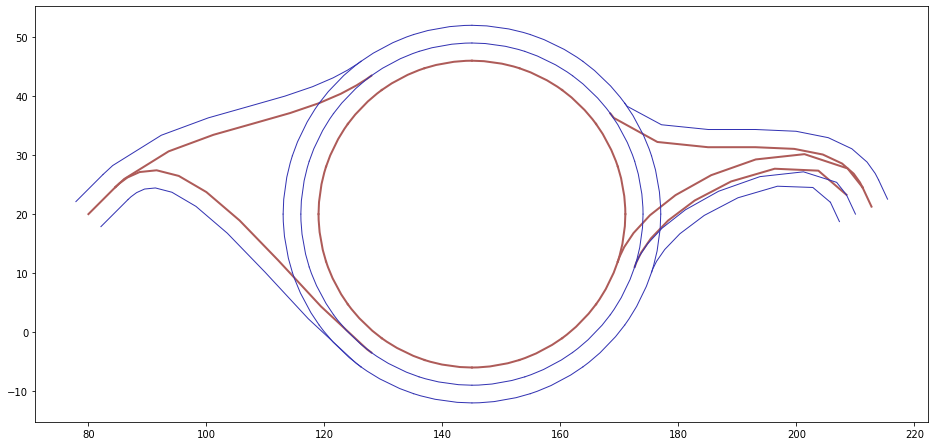}

    \end{subfigure}
    \begin{subfigure}[b]{.35\linewidth}
      \centering
      \includegraphics[width=.95\linewidth]{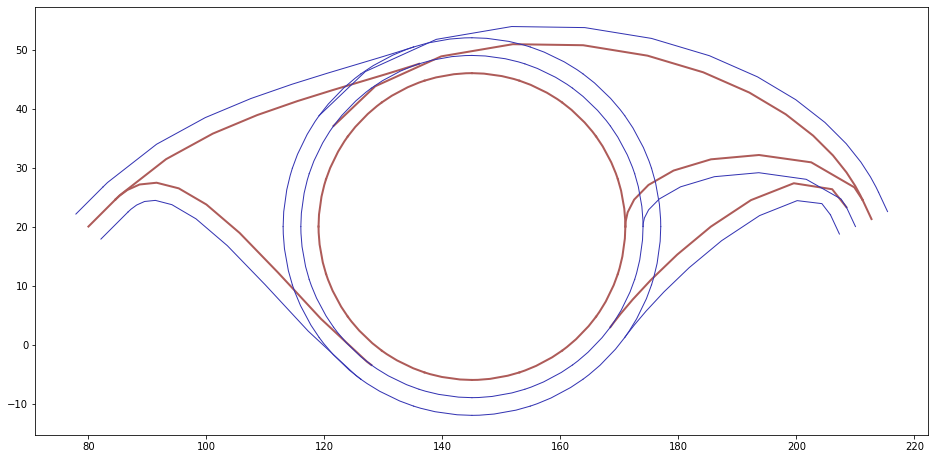}

    \end{subfigure}
    \begin{subfigure}[b]{.35\linewidth}
      \centering
     \includegraphics[width=.95\linewidth]{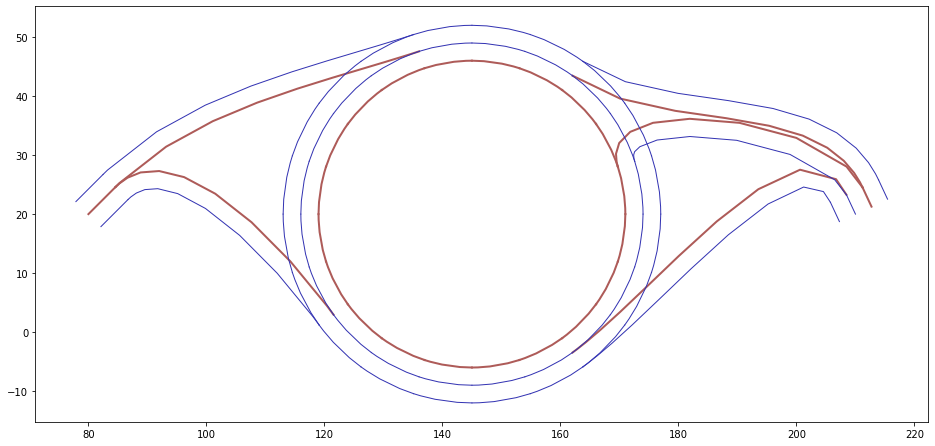}

    \end{subfigure}
    \caption{Diverse roundabouts generated by GFlowNets sampling}.
    \label{fig:fives}
\end{figure}

In this section, we provide other experiments mentioned in Section~4. We study the effect of GFlowNets on two problems: 
\begin{itemize}
    \item 2 incident road roundabouts with a 2-2 lane split ($N$ = 4)
    \item 2 incident road roundabouts with a 2-3 lane split ($N$ = 5)
\end{itemize} 
We pick $C$ = 15 and 20 for the 2-2 and 2-3 split, respectively. We train the GFlowNets for 1000 episodes with a batch size $b$ = 16. For 2-2 split, we sample 2000 configurations after training and find 107 modes. For 2-3 split, we sample 16000 configurations and find 157 distinct modes. It is worth noting that during training, 146 distinct modes were found, and after training, only 9 modes repeat again (Figure \ref{fig:fiveLanes}). Thus, the GFlowNets agent samples configurations not seen before. Figure \ref{fig:fives} shows some of the outputs and the diversity in them.

\section{Algorithms}
\label{app:algorithm}



In this section, we provide the alogithms mentioned from Section~3.3.
\begin{algorithm}[ht!]
    \caption{GFlowNet Training}
    \label{Algo:train}
    \textbf{Input:}\\
    $P_F$: Forward flow \\
    $P_B$: Backward flow (initialized as uniform flow)\\
    $Z_\theta$: Logarithmic total flow learned by neural network\\
    $N$ = Total number of lanes\\
    $m$ = Minibatch size\\
    \For{each episodes}{
        $s \leftarrow \{s_0^1, \dots s_0^m\}$\\
        Initialize trajectory $\mathcal{T}$ to an empty list\\
        \While{$s$ not a terminal state}{
            Sample actions  $A = \{a^1, \dots a^m\}$ based on $P_F(s; \theta)$\\
            $s' \leftarrow transition(s, A)$\\
            Append $(s', A)$ to the $\mathcal{T}$\\
            $s \leftarrow s'$
        }
        
        Compute $R_{proxy}$(s) using reward function using equation \ref{rewardFunction} from $\mathcal{T}$ using the last state in $\mathcal{T}$\\
        Compute loss $\mathcal{L}$ using trajectory balance loss function in  \ref{lossFunction} from $\mathcal{T}$\\
        Update parameters $\theta$ using stochastic descent on loss $\mathcal{L}$\\
    }
\end{algorithm}
\begin{algorithm}[ht!]
    \caption{Roundabout Generation With GFlowNet}
    \label{Algo:generate}
    \textbf{Input: }\\ $P$: A set of road configurations ${(x_i, y_i, h_i, nLeftLanes_i, nRightLanes_i)}$\\
    $N$ = Total number of lanes\\
    $\pi_\theta$: Generative policy learned by GFlowNet trained  using reward function R \\
    $C$ : Number of circular segments\\
    $b$ : Batch size\\
    \KwResult{$O$ : Roundabout-like road geometries having the best rewards}
    \textbf{Algorithm}\\
    Create circular road junction using $C$ and $P$ \\
    Train $\pi_\theta$ using Proxy Reward function $R_{proxy}$\\
    Sample batch $B$ = ${x_i, \dots, x_b}$ with $x_i$ $\sim$ $\pi_\theta$\\
    Extract $top-K$ by taking distinct $K$ configurations from $B$ with highest proxy rewards\\ 
    Connect incident roads using configurations from $BestEncodings$ and build $|BestEncodings|$ roundabouts
\end{algorithm}

\end{document}